\documentclass[lettersize,journal]{IEEEtran}
\usepackage{amsmath,amsfonts}
\usepackage{algorithmic}
\usepackage{algorithm}
\usepackage{array}
\usepackage[caption=false,font=normalsize,labelfont=sf,textfont=sf]{subfig}
\usepackage{textcomp}
\usepackage{stfloats}
\usepackage{url}
\usepackage{verbatim}
\usepackage{graphicx}
\usepackage{longtable}
\usepackage{cite}
\usepackage{cite}
\usepackage{hyperref}
\usepackage{orcidlink}
\hypersetup{
  colorlinks   = true,
  urlcolor     = blue,
  linkcolor    = blue,
  citecolor    = blue
}

\begin{document}

\title{\title{ReDAG$^{\mathrm{\textbf{RT}}}$: Global Rate-Priority Scheduling for Real-Time Multi-DAG Execution in ROS~2}}
\author{Md. Mehedi Hasan\,\orcidlink{0009-0000-1078-5778}, Rafid Mostafiz\,\orcidlink{0000-0002-5905-6530}, Bikash Kumar Paul\,\orcidlink{0000-0002-4414-2751}, Md. Abir Hossain\,\orcidlink{0000-0003-3651-3345}, and Ziaur Rahman\,\orcidlink{0000-0002-7759-3428}  ~\IEEEmembership{}
\thanks{The source code and experimental artifacts of 
ReDAG$^{\mathrm{RT}}$ are openly available at: 
\url{https://github.com/Mehedi16009/ros2-realtime-scheduling-framework}}
\thanks{}
\thanks{}}
\markboth{}%
{Shell \MakeLowercase{\textit{et al.}}: A Sample Article Using IEEEtran.cls for IEEE Journals}


\maketitle

\begin{abstract}
ROS~2 has become the dominant middleware for robotic systems, where perception, estimation, planning, and control pipelines are naturally structured as directed acyclic graphs (DAGs) of callbacks executed under a shared executor. Despite widespread adoption, default ROS~2 executors employ best-effort dispatch with no cross-DAG priority enforcement, producing uncontrolled callback contention, structural priority inversion, and deadline instability under concurrent workloads. These limitations restrict deployment in time-critical and safety-sensitive cyber-physical systems (CPS). This paper presents ReDAG$^{\mathrm{RT}}$, a user-space global scheduling framework that enables deterministic multi-DAG execution within unmodified ROS~2. The core contribution is a Rate-Priority (RP)-driven global ready-queue architecture that orders callbacks system-wide by activation rate, enforces per-DAG concurrency bounds, and mitigates cross-graph priority inversion, without modifying the ROS~2 API, executor interface, or the Linux scheduler. A multi-DAG task model is formalized for ROS~2 callback pipelines, and cross-DAG interference is characterized under RP scheduling. Response-time recurrences and schedulability conditions are derived within the classical Rate-Monotonic theory. Controlled experiments within a ROS~2 Humble execution environment compare ReDAG$^{\mathrm{RT}}$ against the \texttt{SingleThreadedExecutor} and \texttt{MultiThreadedExecutor} using synthetic multi-DAG workloads under mixed-period, mixed-criticality configurations. Results show up to $29.7\%$ reduction in combined deadline miss rate, $42.9\%$ compression of the $99$th-percentile response time, and a $13.7\%$ improvement in miss rate relative to the 
\texttt{MultiThreadedExecutor} under comparable utilization. Asymmetric per-DAG concurrency bounds further reduce interference by $40.8\%$ over symmetric configurations. These results demonstrate that deterministic and analyzable multi-DAG scheduling is achievable entirely within the ROS~2 user-space execution layer, providing a principled foundation for real-time robotic middleware in safety-critical cyber-physical deployments.
\end{abstract}


\section{Introduction}
\begin{figure}[!t]
\centering
\includegraphics[width=3.65in]{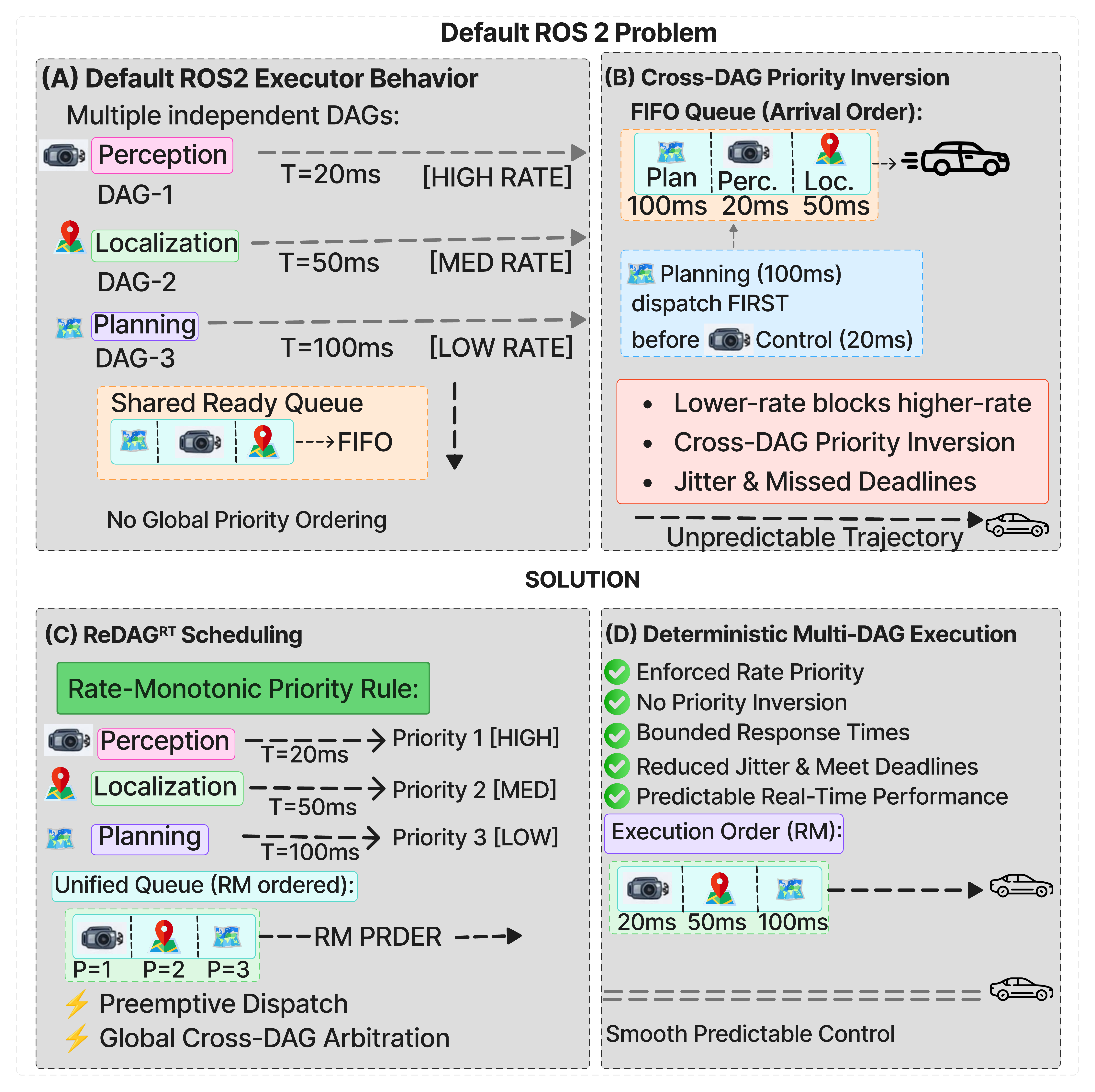}
\caption{Cross-DAG priority inversion under the default ROS~2 executor and the ReDAG$^{\mathrm{RT}}$ solution. (A)~Multiple independent DAGs collapse into a shared FIFO queue with no global priority ordering. (B)~Low-frequency callbacks execute before high-frequency ones, causing priority inversion and deadline misses. (C)~ReDAG$^{\mathrm{RT}}$ enforces Rate-Monotonic ordering via a unified priority queue. (D)~Deterministic execution eliminates priority inversion and bounds response times.}
\label{Fig_1}
\end{figure}
\IEEEPARstart{M}{odern} robotic systems deploy multiple concurrent computation pipelines within a shared middleware runtime. Perception, localization, planning, and control are commonly structured as independent directed acyclic graphs (DAGs) of callbacks, all executed under a unified executor. In ROS~2, these callbacks are managed by either a \texttt{SingleThreadedExecutor} or a \texttt{MultiThreadedExecutor}, each maintaining a global ready queue \cite{R1}. While this model simplifies concurrency management, it imposes no global real-time scheduling policy across application DAGs\cite{R2}. The default executor operates at the callback level without enforcing rate-based priority relations. A callback becomes eligible for execution when its triggering condition is satisfied and is then dispatched according to queue ordering and thread availability. No mechanism ensures that higher-rate callbacks dominate lower-rate ones across distinct DAGs\cite{R3,R4}. Consequently, execution order is determined by arrival timing and worker scheduling rather than real-time priority semantics. This design induces cross-DAG interference. A callback belonging to a low-frequency pipeline may execute before a high-frequency control callback if it enters the ready queue earlier or acquires a worker thread first, a phenomenon we identify as \emph{cross-DAG priority inversion}~\cite{R5,R6}, as illustrated in Fig.~\ref{Fig_1}. Multiple application DAGs structurally collapse into a shared callback queue under the default executor, producing priority inversion across otherwise independent pipelines. The resulting interference increases response-time variability and degrades temporal predictability. Under moderate to high processor utilization, these effects amplify jitter and cause deadline misses. From a real-time systems perspective, this behavior directly violates classical Rate-Monotonic scheduling principles~\cite{R7, R8}. Although individual DAGs may internally respect precedence constraints, the executor layer collapses all graphs into a shared queue without preserving a global priority order\cite{R9,R10}. Default ROS~2 execution models therefore lack a schedulability-aware abstraction that spans multiple concurrent DAGs within a single process. The core architectural insight of ReDAG$^{\mathrm{RT}}$ is that cross-DAG interference originates from the absence of a unified rate-priority hierarchy at the executor level. Mapping all callbacks to globally ordered schedulable entities restores the applicability of classical fixed-priority analysis across the entire callback set. Enforcing Rate-Monotonic priority assignment across DAG boundaries eliminates structural priority inversion and recovers analyzable timing behavior\cite{R11}.
 
ReDAG$^{\mathrm{RT}}$ addresses this limitation by introducing a global rate-priority scheduler integrated directly within the ROS~2 execution layer. All ready callbacks are inserted into a centralized priority queue ordered by activation period, and dispatch decisions follow fixed-priority real-time semantics. Intra-DAG precedence constraints are preserved, while cross-DAG execution is globally arbitrated according to Rate-Monotonic ordering. This design transforms the default timing-agnostic executor into a schedulability-aware runtime without modifying the ROS~2 API, application-level graph structure, or the underlying operating system scheduler. By restoring deterministic priority ordering across heterogeneous callback pipelines, ReDAG$^{\mathrm{RT}}$ bounds cross-graph interference and stabilizes end-to-end response times, enabling principled real-time reasoning for multi-DAG robotic workloads executing within a shared middleware process.

\smallskip
\noindent\textbf{Contributions.} This paper makes three primary contributions:
\begin{itemize}
    \item \textbf{Structural analysis of ROS~2 executor limitations.} Cross-DAG priority inversion is identified and characterized as a fundamental structural deficiency of the default ROS~2 execution model, originating from the absence of global priority arbitration across concurrently executing callback pipelines.

    \item \textbf{Formal interference model for multi-DAG scheduling.} The absence of rate-priority enforcement is shown to produce uncontrolled cross-DAG 
    interference, timing nondeterminism, and degraded deadline predictability under concurrent workloads. Interference bounds and response-time recurrences are derived within a formal multi-DAG task model grounded in classical Rate-Monotonic theory.

    \item \textbf{Design, evaluation, and open-source release of ReDAG$^{\mathrm{RT}}$.} A unified Rate-Monotonic scheduling framework is presented that introduces a global ready-queue executor enforcing deterministic multi-DAG callback arbitration entirely within the ROS~2 user-space execution 
    layer. The framework requires no kernel modifications, preserves full ROS~2 API compatibility, and is empirically validated under mixed-period, mixed-criticality workloads against standard ROS~2 executors. The full implementation is openly released to facilitate reproducibility and community adoption.
\end{itemize}

The remainder of this paper is organized as follows. Section~\ref{sec:related} reviews related work and positions ReDAG$^{\mathrm{RT}}$ against existing ROS~2 real-time scheduling approaches. Building on this, Section~\ref{sec:methodology} presents the system design, formal task model, rate-priority scheduling framework, interference characterization, and response-time analysis. Following the methodology, Section~\ref{sec:results} presents an empirical evaluation of thread scalability, deadline sensitivity, and cross-system comparisons. The subsequent Section~\ref{sec:discussion} provides a scheduling-theoretic interpretation of the results. Finally, Section~\ref{sec:conclusion} concludes and outlines future directions.

\section{RELATED WORK}
\label{sec:related}
\subsection{Real-Time Scheduling Foundations}
Classical real-time scheduling theory establishes the analytical basis for predictable task execution in safety-critical systems. Liu and Layland introduced Rate-Monotonic (RM) and Earliest-Deadline-First (EDF) scheduling for periodic task systems, along with utilization bounds and response-time analysis for fixed-priority preemption \cite{R12}. Subsequent work extended these results to Directed Acyclic Graph (DAG) task models, characterizing intra-task parallelism, precedence constraints, and interference under multiprocessor scheduling \cite{R13}. These models formalize the conditions under which bounded response times and deadline adherence can be guaranteed. In contrast, ROS~2’s default \texttt{SingleThreadedExecutor} and \texttt{MultiThreadedExecutor} employ best-effort FIFO callback dispatch without global priority arbitration across independent DAG pipelines. As a result, classical schedulability assumptions do not directly hold in practical robotic middleware deployments, motivating structured user-space scheduling mechanisms.

\subsection{ROS 2 Real-Time Evolution}
Multiple studies have examined timing behavior in ROS~2. Casini et al.\cite{R14} demonstrate that executor processing windows and callback ordering can introduce unfair dispatch and increased end-to-end latency in chained processing pipelines \cite{R11}. Macenski et al.\cite{R15} describe ROS~2’s production-oriented redesign with DDS-based QoS controls, while acknowledging that executor-level scheduling remains largely FIFO-driven. Kernel-level interventions such as \texttt{PREEMPT\_RT} improve worst-case interrupt latency but require system-level modifications and do not address callback-level arbitration within the executor \cite{R16}. Similarly, improvements in Linux scheduling fairness may enhance throughput but do not establish analyzable dispatch order among competing callbacks \cite{R16}. These efforts improve system-level timing characteristics yet leave executor-level global prioritization largely unaddressed.

\subsection{Chain-Aware Scheduling Approaches}
Chain-aware scheduling represents a partial solution to timing predictability. PiCAS assigns priorities to callbacks according to end-to-end chain deadlines and demonstrates improved latency bounds in embedded robotic platforms \cite{R17,R18}. However, its analysis primarily considers isolated chains and does not explicitly characterize interference among heterogeneous DAG pipelines. RAGE adopts chain-based prioritization but lacks formal schedulability analysis and global queue arbitration \cite{R19}. PoDS introduces priority inheritance mechanisms for micro-ROS environments \cite{R20}, focusing on MCU-scale deployments rather than multi-core robotic workloads. While these works strengthen intra-chain timing guarantees, cross-DAG contention in shared executors remains insufficiently characterized.

\subsection{Executor-Level Scheduling Extensions}
Several efforts modify ROS~2 executors to better align with classical scheduling models. Teper et al.\ and Bell et al.\ extend the \texttt{EventsExecutor} to support RM/EDF-compatible dispatch and enable response-time analysis under constrained triggering models \cite{R21,R22}. These approaches rely on specific DDS configurations and restricted callback models, limiting applicability to general robotic pipelines with mixed subscription and timer callbacks. ROSRT leverages Linux \texttt{SCHED\_EXT} to provide flexible thread-level preemption without kernel patches \cite{R23}. However, scheduling decisions remain tied to DDS-topic granularity and do not introduce unified global arbitration across multiple DAGs. ROSCH synchronizes DAG timestamps for Autoware-based systems \cite{R24}, yet remains application-specific and conservative in delay estimation.

\subsection{Multi-DAG Scheduling and Analysis Gaps}
\label{sec:Multi-DAG Scheduling and Analysis Gaps}
Theoretical multi-DAG scheduling has been studied in broader real-time systems literature. Yano et al.\ apply GEDF to Autoware sub-DAGs but assume FIFO-ready queues inconsistent with asynchronous ROS~2 callback behavior \cite{R25}. Ahmed proposes an rp-sporadic DAG scheduler with self-dependency handling \cite{R26}, though without integration into ROS middleware. Tang et al.\ derive response-time bounds under reservation-based chain models \cite{R27}, primarily focusing on single-chain interference. Blass analyzes callback interference in the \texttt{MultiThreadedExecutor} \cite{R28}, yet does not introduce priority-driven global queue arbitration. Chaaban et al.\ propose executor grouping heuristics to reduce resource contention \cite{R29}, but without formal multi-DAG schedulability characterization. Collectively, existing work either (i) prioritizes single chains, (ii) modifies executor backends under constrained assumptions, or (iii) analyzes interference without introducing structured global arbitration across heterogeneous DAG pipelines.

\subsection{Positioning Against Prior Art}
Table~\ref{tab:lit-review} summarizes representative approaches. Default executors provide compatibility but no global priority ordering. Chain-aware schedulers improve end-to-end latency within individual pipelines. EventsExecutor modifications enable classical analysis under restricted triggering assumptions. OS-level approaches enhance thread scheduling but do not directly address callback-level multi-DAG arbitration.

\begin{table*}[t]
\centering
\caption{Comparison of ROS 2 Real-Time Scheduling Approaches}
\label{tab:lit-review}
\begin{tabular}{lccccccc}
\hline
Approach & Multi-DAG & Global Queue & Kernel Mod & API Compat & Formal Analysis & Evaluation \\ \hline
Default Execs \cite{R11} & $\times$ & $\times$ & $\checkmark$ & $\checkmark$ & $\times$ & Baseline \\
PiCAS \cite{R17,R18} & $\times$ & $\times$ & ? & ? & Chain-level & Real HW \\
EventsEx \cite{R21,R22} & Tree & $\times$ & $\checkmark$ & $\sim$ & RTA (restricted) & Sim \\
ROSRT \cite{R23} & $\checkmark$ & OS-level & $\checkmark$ & $\checkmark$ & $\times$ & Autoware \\
ROSCH \cite{R24} & $\checkmark$ & $\times$ & ? & $\sim$ & DAG-sync & Autoware \\
PREEMPT\_RT \cite{R16} & $\checkmark$ & Linux & $\times$ & $\checkmark$ & $\times$ & Throughput \\ \hline
\textbf{ReDAG$^{\mathrm{RT}}$} & $\checkmark$ & $\checkmark$ & $\checkmark$ & $\checkmark$ & Multi-DAG RTA & Webots + Stress \\ \hline
\end{tabular}
\end{table*}

\subsection{Research Gap and Motivation}
To our knowledge, no prior work provides user-space global rate-priority queue arbitration across heterogeneous DAG pipelines within unmodified ROS~2 executors while enabling multi-DAG response-time and lateness analysis. Existing approaches either restrict the callback model, depend on specialized executor variants, rely on OS-level scheduling, or focus on single-chain optimization. ReDAG$^{\mathrm{RT}}$ addresses this gap by introducing structured global rate-priority dispatch within the standard executor abstraction, preserving ROS~2 API compatibility and avoiding kernel modifications. Its evaluation explicitly examines cross-DAG interference, response-time variability, and deadline behavior under stress workloads, thereby complementing prior single-chain and OS-centric approaches with a middleware-level multi-DAG scheduling framework.


\section{ReDAG$^{\mathrm{RT}}$: System Design and Scheduling Framework}
\label{sec:methodology}
\subsection{System Architecture Overview}
\label{sec:System Architecture Overview}
\begin{figure}[!t]
\centering
\includegraphics[width=3.65in]{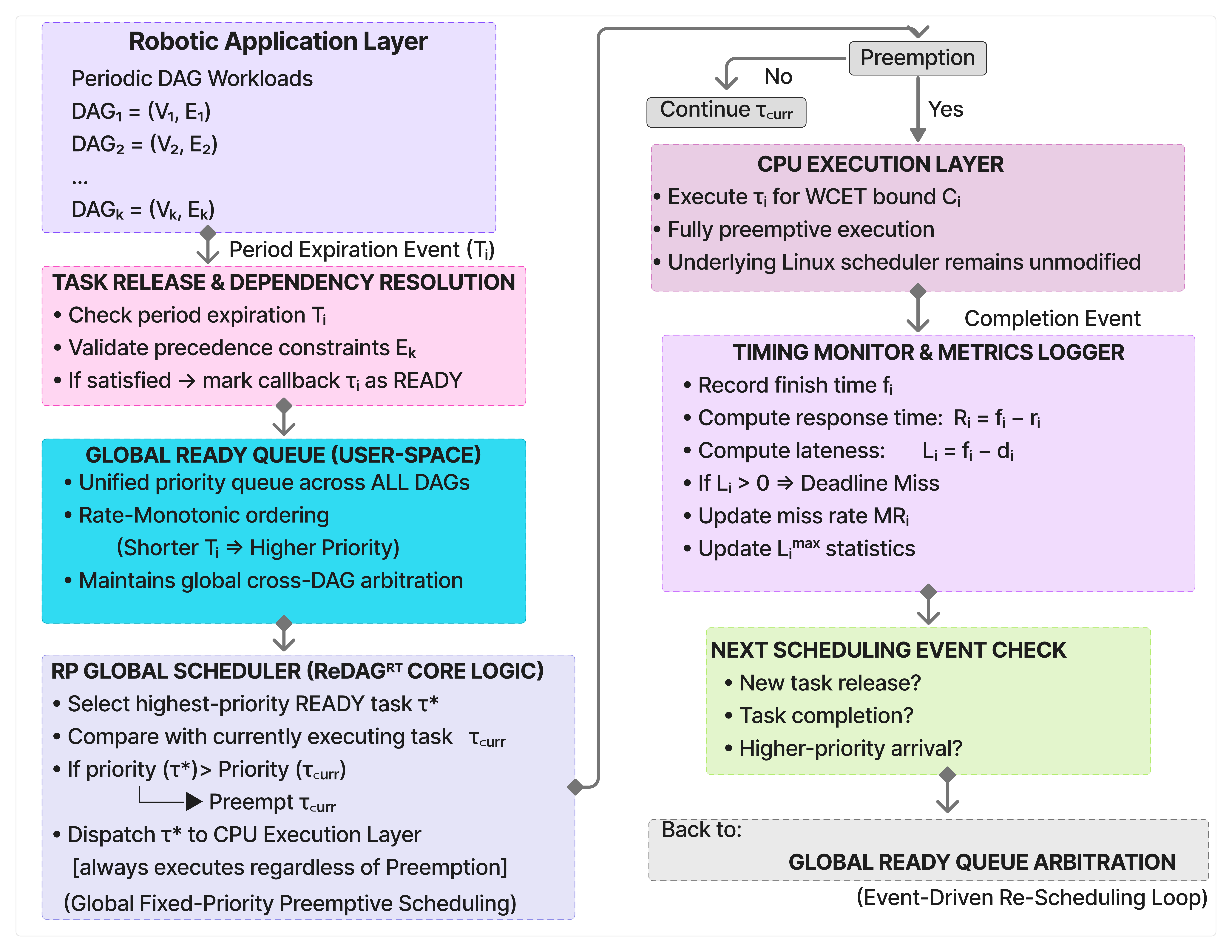}
\caption{ReDAG$^{\mathrm{RT}}$ layered user-space scheduling architecture. Periodic DAG workloads $G_k = (V_k, E_k)$ release tasks upon period expiration. A dependency-aware release stage validates timing and precedence constraints before admitting ready tasks into a unified global Rate-Monotonic priority queue ordered by $\arg\min_{\tau_i \in \mathcal{R}}\,T_i$. The ReDAG$^{\mathrm{RT}}$ executor performs fixed-priority preemptive 
arbitration across all DAGs and dispatches to the unmodified Linux execution layer. An integrated Timing Monitor computes $R_i$, $L_i$, and $\mathrm{MR}_i$ online, with event-driven re-scheduling closing the control loop.}
\label{Fig_2}
\end{figure}

ReDAG$^{\mathrm{RT}}$ is implemented as a custom global executor within the ROS~2 user-space execution layer and is structured into three logical tiers: (i) an Application Layer hosting concurrent DAG-structured robotic pipelines, (ii) an Executor Layer enforcing deterministic Rate-Priority (RP) scheduling through a unified global ready queue, and (iii) an OS Layer comprising the standard Linux scheduler operating without kernel modification. The complete architecture and event-driven execution flow are illustrated in~\hyperref[Fig_2]{Fig.~2}. At the application layer, the system executes multiple concurrent directed acyclic graphs $G_k = (V_k, E_k)$, where $V_k$ denotes the set of callback tasks and $E_k$ encodes inter-task precedence constraints. Each DAG represents an independent robotic pipeline, such as perception, estimation, planning, or control. Tasks are released periodically according to their specified periods $T_i$, consistent with implicit-deadline periodic task models. This layer remains fully compatible with standard ROS~2 APIs and does not require application-level modifications. Upon period expiration, a release event triggers the Task Release and Dependency Resolution stage. For each task $\tau_i$, the resolver verifies two conditions: (1) that its period $T_i$ has elapsed, and (2) that all predecessor tasks specified in $E_k$ have completed. Tasks satisfying both conditions are marked \textsc{ready} and inserted into the Global Ready Queue. This queue is a unified user-space priority structure shared across all concurrent DAGs. The queue enforces Rate-Monotonic priority ordering, assigning strictly higher priority to tasks with shorter periods (i.e., $T_i < T_j \Rightarrow P_i > P_j$). This establishes a single system-wide arbitration order across all DAGs, eliminating per-DAG scheduling inconsistencies and uncontrolled cross-DAG interference observed in the default ROS~2 {\tt SingleThreadedExecutor} and {\tt MultiThreadedExecutor}, which rely on FIFO or loosely ordered callback dispatching without global priority guarantees. The RP Global Scheduler constitutes the core of ReDAG$^{\mathrm{RT}}$ and implements fixed-priority preemptive scheduling entirely in user space. At each scheduling point, the scheduler selects the highest-priority task from the ready queue and compares it with the currently executing task. If the selected task has strictly higher priority, preemption is triggered. The selected task $\tau_i$ is then dispatched to the CPU execution layer. During execution, each task runs for at most its worst-case execution time $C_i$ under fully preemptive semantics. The underlying Linux thread scheduler remains unmodified; ReDAG$^{\mathrm{RT}}$ controls dispatch ordering exclusively at the executor level and does not alter kernel-level scheduling policies. Upon completion, a completion event is forwarded to the Timing Monitor and Metrics Logger. The finish time $f_i$ is recorded, and timing metrics are computed as follows. The response time is defined as $R_i = f_i - r_i$, where $r_i$ denotes the release time. Lateness is defined as $L_i = f_i - d_i$, where $d_i$ represents the absolute deadline. A deadline miss is registered whenever $L_i > 0$, and the corresponding miss counter is incremented. Control then transitions to the Next Scheduling Event Check, which determines whether the triggering event corresponds to a new task release, a task completion, or the arrival of a higher-priority task. In all cases, control returns to the Global Ready Queue Arbitration stage, forming a closed event-driven scheduling loop. This architecture ensures continuous, globally prioritized execution across all concurrent DAGs while enabling precise per-task timing analysis and systematic metric collection under user-space constraints.

\subsection{Multi-DAG Task Model Formulation}
The computational workload of ReDAG$^{\mathrm{RT}}$ is modeled as a collection of directed acyclic graphs (DAGs), where each DAG captures the structural and temporal dependencies of an independent robotic processing pipeline. Formally, each DAG instance is defined as:

\begin{equation}
    G_k = (V_k, E_k),
    \label{eq:dag}
\end{equation}

\noindent where $V_k = \{\tau_1, \tau_2, \dots, \tau_n\}$ denotes the set of callback tasks belonging to $G_k$, and $E_k \subseteq V_k \times V_k$ represents the set of directed precedence constraints between tasks. An edge $(\tau_i, \tau_j) \in E_k$ indicates that task $\tau_j$ cannot be released for execution until task $\tau_i$ has completed. Each task $\tau_i \in V_k$ corresponds to a ROS~2 callback, such as a timer callback, subscription callback, or computation stage within a robotic pipeline. The acyclic property of $G_k$ guarantees the absence of circular dependencies, thereby ensuring well-defined execution orderings within each pipeline. The system hosts multiple such DAGs executing concurrently. The global workload is defined as:

\begin{equation}
    \mathcal{G} = \{G_1, G_2, \dots, G_m\},
    \label{eq:global_workload}
\end{equation}

\noindent where each $G_k$ models an independent robotic application component, as 
formalized in Equation~\eqref{eq:dag}. The DAGs in $\mathcal{G}$ are assumed to be 
functionally independent in terms of logical correctness; that is, no precedence 
constraints exist across distinct DAGs. However, all DAGs share a common execution 
environment and compete for the same executor resources within ReDAG$^{\mathrm{RT}}$. This shared-resource assumption introduces cross-DAG contention at the scheduling level, which directly motivates the need for a unified global arbitration mechanism. By formally separating structural independence at the application level from resource sharing at the executor level, the model in Equations~\eqref{eq:dag} 
and~\eqref{eq:global_workload} captures the realistic deployment scenario in which multiple robotic pipelines operate concurrently while relying on a single scheduling substrate.

\subsection*{B.1 Task Characterization and Timing Parameters}
Each task $\tau_i$ in the multi-DAG system is characterized by a tuple of timing 
and criticality parameters:
\begin{equation}
    \tau_i = (C_i,\ T_i,\ D_i,\ \chi_i),
    \label{eq:task_tuple}
\end{equation}

\noindent where $C_i$ denotes the worst-case execution time (WCET), $T_i$ represents 
the task period, $D_i$ is the relative deadline, and $\chi_i$ specifies the 
criticality level of the task. This formulation captures both temporal constraints 
and system-level importance within a unified abstraction. The WCET $C_i$ defines an upper bound on the processor time required by $\tau_i$ under worst-case conditions, assumed to be known a priori through measurement, static analysis, or conservative estimation. The period $T_i$ determines the minimum inter-arrival time between consecutive releases of the same task, reflecting the sensing or control frequency of the robotic application. The relative deadline $D_i$ constrains the maximum allowable response time from task release to completion. The criticality level $\chi_i$ enables differentiation between safety-critical and best-effort callbacks, providing a foundation for future extensions toward mixed-criticality scheduling. Throughout this work, implicit deadlines are assumed for all tasks:

\begin{equation}
    D_i = T_i.
    \label{eq:implicit_deadline}
\end{equation}

\noindent This assumption, defined in Equation~\eqref{eq:implicit_deadline}, aligns each task's completion requirement with the arrival of its next instance, simplifying schedulability analysis and remaining consistent with periodic control and perception workloads common in robotic systems. Let $N$ denote the total number of tasks across all DAGs in $\mathcal{G}$, as defined in Equation~\eqref{eq:global_workload}. The overall processor utilization of the system is defined as:

\begin{equation}
    U = \sum_{i=1}^{N} \frac{C_i}{T_i}.
    \label{eq:utilization}
\end{equation}

\noindent This utilization metric, given in Equation~\eqref{eq:utilization}, provides a compact measure of aggregate computational demand relative to processor capacity. It serves as a primary indicator of schedulability under rate-based global arbitration and enables analytical reasoning about system feasibility when multiple DAGs defined in Equation~\eqref{eq:dag} share the same executor resources within ReDAG$^{\mathrm{RT}}$.

\subsection{Rate-Priority Global Scheduling Design}
\label{sec:Rate-Priority Global Scheduling Design}
ReDAG$^{\mathrm{RT}}$ employs a static Rate-Monotonic priority assignment enforced globally across all DAGs, transforming cross-DAG callback arbitration into a formally analyzable fixed-priority scheduling problem.
\subsection*{C.1 Static Priority Assignment}
ReDAG$^{\mathrm{RT}}$ adopts a fixed-priority scheduling policy based on Rate-Monotonic (RM) ordering. Each task $\tau_i$ is assigned a static priority determined solely by its period, according to the relation:

\begin{equation}
    T_i < T_j \implies P_i > P_j,
    \label{eq:rm_priority}
\end{equation}

\noindent such that tasks with shorter periods receive strictly higher priorities. This mapping reflects the principle that higher-frequency tasks impose tighter timing requirements and therefore demand stronger scheduling guarantees. Priority assignment is performed globally across the entire task set $\mathcal{G}$, 
defined in Equation~\eqref{eq:global_workload}, rather than locally within individual DAGs. All tasks from all graphs are placed into a unified priority space, allowing a high-rate task in one DAG to preempt a lower-rate task in another. This global ordering, formalized in Equation~\eqref{eq:rm_priority}, ensures consistent 
arbitration under shared executor resources and prevents priority inversion across independent robotic pipelines. Furthermore, the static nature of the assignment eliminates runtime priority recomputation, reduces scheduling overhead, and enables analytical reasoning about worst-case interference.

\subsection*{C.2 Schedulability Bound Under Rate-Monotonic Policy}
For $N$ independent periodic tasks scheduled under the classical Rate-Monotonic policy on a single processor, schedulability is guaranteed if the total processor utilization $U$, defined in Equation~\eqref{eq:utilization}, satisfies:

\begin{equation}
    U \leq N\left(2^{1/N} - 1\right).
    \label{eq:rm_bound}
\end{equation}

\noindent This bound represents a sufficient condition derived from worst-case response-time analysis. As $N$ increases, the right-hand side of Equation~\eqref{eq:rm_bound} converges to $\ln(2) \approx 0.693$, providing a conservative analytical reference for system design. In the context of ReDAG$^{\mathrm{RT}}$, tasks are not fully independent due to intra-DAG precedence constraints encoded in $E_k$, as defined in Equation~\eqref{eq:dag}. These constraints restrict the set of feasible release patterns and can, in practice, reduce inter-task interference relative to the fully independent case. Nevertheless, the classical RM utilization bound in Equation~\eqref{eq:rm_bound} is retained as a conservative feasibility guideline. When the total system utilization $U$ satisfies this condition, the system operates within a known safe analytical region, even before accounting for the structural restrictions imposed by the DAG topology.

\subsection{Ready-Queue Management and Dispatch Semantics}
ReDAG$^{\mathrm{RT}}$ employs a single global ready queue containing all tasks eligible for execution across every DAG in $\mathcal{G}$, as defined in Equation~\eqref{eq:global_workload}. The queue is strictly ordered according to the static Rate-Monotonic priority assignment established in Equation~\eqref{eq:rm_priority}, such that tasks with shorter periods are placed ahead of tasks with longer periods, ensuring deterministic global arbitration across all concurrent pipelines. A task $\tau_i$ becomes eligible for insertion into the ready queue only when two conditions are simultaneously satisfied: its release time has arrived and all intra-DAG precedence constraints $E_k$, defined in Equation~\eqref{eq:dag}, have been resolved. Once both timing and dependency conditions hold, the corresponding callback is marked \textsc{ready} and admitted into the global priority queue. At each scheduling point, triggered by a task release, task completion, or preemption event, the scheduler performs three sequential actions: all newly eligible callbacks are enqueued, the highest-priority task is identified from the ready set, and preemption is triggered if that task holds strictly higher priority than the currently executing task. Formally, the dispatch decision is defined as:

\begin{equation}
    \tau^* = \arg\min_{i \in \mathcal{R}}\ T_i,
    \label{eq:dispatch}
\end{equation}

\noindent where $\mathcal{R}$ denotes the set of currently ready tasks. The scheduler selects the task with the minimum period from $\mathcal{R}$, which corresponds directly to the highest Rate-Monotonic priority under the assignment in Equation~\eqref{eq:rm_priority}. This dispatch rule ensures that the processor is always allocated to the most time-critical eligible task, consistent with the utilization bound established in Equation~\eqref{eq:rm_bound}. Deadline monitoring is performed upon task completion. When $\tau_i$ finishes execution, its completion time $f_i$ is recorded and evaluated against its absolute deadline $d_i = r_i + D_i$, where $D_i = T_i$ under the implicit deadline assumption of Equation~\eqref{eq:implicit_deadline}. Lateness $L_i = f_i - d_i$ is computed and a deadline miss is registered if $L_i > 0$, as defined in the timing model of \hyperref[sec:System Architecture Overview]{Section~\ref*{sec:System Architecture Overview}}. This design keeps dispatch logic lightweight while ensuring that timing violations are detected precisely at job termination.

\subsection{ReDAG\texorpdfstring{$^{\mathrm{RT}}$}{RT} Executor Algorithm}
The complete runtime behavior of ReDAG$^{\mathrm{RT}}$ is formalized in Algorithm~\ref{alg:redag_rt}. The algorithm integrates periodic release handling, precedence verification, global Rate-Priority arbitration, preemptive dispatch, and deadline monitoring within a unified execution loop, representing the concrete realization of the scheduling semantics established in Equations~\eqref{eq:rm_priority}--\eqref{eq:dispatch} and the interference model of Equation~\eqref{eq:interference}. At each iteration, the algorithm processes all tasks $\tau_i \in \mathcal{G}$ whose period $T_i$ has expired. For each such task, the dependency resolver evaluates the intra-DAG precedence constraints $E_k$, as defined in Equation~\eqref{eq:dag}, confirming that all predecessor tasks have completed. Tasks satisfying both the temporal and precedence conditions are marked \textsc{ready} and inserted into the global priority queue ordered by Equation~\eqref{eq:rm_priority}. The dispatcher then applies the selection rule of Equation~\eqref{eq:dispatch} to identify the highest-priority ready task $\tau^* = \arg\min_{i \in \mathcal{R}}\ T_i$, triggering preemption if $\tau^*$ holds strictly higher priority than the currently executing task. Upon task completion, the Timing Monitor records $f_i^k$ and evaluates lateness $L_i^k = f_i^k - d_i^k$ per Equation~\eqref{eq:lateness}, registering a deadline miss and updating $\mathrm{MR}_i$ per Equation~\eqref{eq:miss_rate} whenever $L_i^k > 0$. The loop then advances to the next scheduling event, maintaining the closed-loop arbitration structure described in \hyperref[sec:Execution Workflow]{Section~\ref*{sec:Execution Workflow}} until system termination.

\subsection{Interference Characterization}
\label{sec:Interference Characterization}
Cross-DAG interference is analyzed under both default ROS~2 executors and ReDAG$^{\mathrm{RT}}$, establishing the structural conditions under which interference becomes bounded and analytically tractable.
\subsection*{F.1 Interference Under Default ROS~2 Executors}
Under the default ROS~2 executor policies, callback scheduling is driven by FIFO semantics and internal queue behavior rather than formal priority rules, introducing several sources of timing unpredictability. First, FIFO-induced priority inversion occurs when lower-urgency callbacks are serviced ahead of time-sensitive ones solely because they arrived earlier; since no global rate-based ordering is enforced, execution order does not necessarily reflect temporal criticality. Second, cross-DAG blocking is effectively unbounded:

\begin{algorithm}[H]
\caption{ReDAG$^{\mathrm{RT}}$: Rate-Priority Multi-DAG Executor}
\label{alg:redag_rt}
\begin{algorithmic}[1]
\REQUIRE DAG set $\mathcal{G} = \{G_1, G_2, \dots, G_K\}$, 
         each $G_k = (V_k, E_k)$, task parameters 
         $(C_i, T_i, D_i, \chi_i)$ for each $\tau_i \in V_k$
\STATE Initialize Global Ready Queue $RQ \gets \emptyset$
\STATE Initialize $\tau_{\text{curr}} \gets \texttt{NULL}$
\STATE Initialize metrics logger: 
       $L_i^{\max} \gets 0$, miss\_count$_i \gets 0$, 
       total\_jobs$_i \gets 0$ for all $\tau_i$
\WHILE{system is running}
    \FOR{each DAG $G_k \in \mathcal{G}$}
        \FOR{each task $\tau_i \in V_k$ with period $T_i$ expired}
            \IF{all predecessors of $\tau_i$ in $E_k$ completed}
                \STATE Record release time $r_i \gets$ current\_time
                \STATE Increment total\_jobs$_i$
                \STATE Mark $\tau_i$ as \textsc{ready}
                \STATE Insert $\tau_i$ into $RQ$ ordered by 
                       $\arg\min T_i$ (Rate-Monotonic)
            \ENDIF
        \ENDFOR
    \ENDFOR
    \IF{$RQ \neq \emptyset$}
        \STATE $\tau_{\text{next}} \gets \arg\min_{\tau_i \in RQ} T_i$
        \IF{$\tau_{\text{curr}} = \texttt{NULL}$}
            \STATE Dispatch $\tau_{\text{next}}$ on CPU
            \STATE $\tau_{\text{curr}} \gets \tau_{\text{next}}$
        \ELSIF{priority($\tau_{\text{next}}$) $>$ 
               priority($\tau_{\text{curr}}$)}
            \STATE Preempt $\tau_{\text{curr}}$
            \STATE Dispatch $\tau_{\text{next}}$ on CPU
            \STATE $\tau_{\text{curr}} \gets \tau_{\text{next}}$
        \ELSE
            \STATE Continue $\tau_{\text{curr}}$
        \ENDIF
    \ENDIF
    \FOR{each task $\tau_i$ completed during this cycle}
        \STATE Record finish time $f_i \gets$ current\_time
        \STATE Compute $R_i \gets f_i - r_i$
        \STATE Compute $d_i \gets r_i + D_i$
        \STATE Compute $L_i \gets f_i - d_i$
        \IF{$L_i > 0$}
            \STATE Increment miss\_count$_i$
            \STATE $L_i^{\max} \gets \max(L_i^{\max},\ L_i)$
            \STATE $\mathrm{MR}_i \gets 
                   \frac{\text{miss\_count}_i}
                        {\text{total\_jobs}_i}$
        \ENDIF
        \STATE $\tau_{\text{curr}} \gets \texttt{NULL}$
    \ENDFOR
\ENDWHILE
\end{algorithmic}
\end{algorithm}

 callbacks belonging to different logical pipelines share the same executor threads without structured arbitration, allowing a computationally intensive callback in one DAG to delay unrelated callbacks in another DAG with no analytical upper bound on the resulting interference. Third, because scheduling decisions are not derived from explicit timing parameters such as $T_i$ or $C_i$ defined in Equation~\eqref{eq:task_tuple}, worst-case interference cannot be expressed in closed form. Consequently, no formal priority guarantees or schedulability bounds can be established under the default execution model.

\subsection*{F.2 Controlled Interference Under ReDAG\texorpdfstring{$^{\mathrm{RT}}$}{RT}}
ReDAG$^{\mathrm{RT}}$ restructures interference through strict global Rate-Priority ordering. Since priorities are statically assigned according to Equation~\eqref{eq:rm_priority} and globally enforced via the dispatch rule of Equation~\eqref{eq:dispatch}, a task $\tau_i$ can be delayed exclusively by tasks with strictly higher priority. Let $hp(i)$ denote the set of tasks with higher priority than $\tau_i$. The worst-case interference experienced by $\tau_i$ within a scheduling window is bounded by:

\begin{equation}
    I_i = \sum_{\tau_j \in hp(i)} C_j,
    \label{eq:interference}
\end{equation}

\noindent where $C_j$ is the WCET of each higher-priority task as defined in Equation~\eqref{eq:task_tuple}. This expression captures the cumulative execution demand of all tasks in $hp(i)$ that may execute before or preempt $\tau_i$. Lower-priority tasks contribute no interference under the preemptive scheduling semantics enforced by Equation~\eqref{eq:dispatch}. Critically, cross-DAG interference becomes analytically tractable under ReDAG$^{\mathrm{RT}}$. Although tasks originate from structurally independent DAGs in $\mathcal{G}$, as defined in Equation~\eqref{eq:global_workload}, interference is determined solely by global priority ordering rather than graph membership. This transformation converts the previously unbounded and opaque contention of the default executor into a bounded and analyzable quantity expressed in Equation~\eqref{eq:interference}, enabling formal reasoning about response time and schedulability across the full multi-DAG robotic workload.

\subsection{Response-Time and Lateness Analysis}
\label{sec:Response-Time and Lateness Analysis}
Formal worst-case response-time bounds and empirical lateness metrics are derived to provide both analytical schedulability guarantees and measurement-based validation of timing behavior under ReDAG$^{\mathrm{RT}}$.
\subsection*{G.1 Response-Time Recurrence}
To establish temporal guarantees under global Rate-Priority scheduling, the worst-case response time of each task $\tau_i$ is computed using fixed-point iteration. The response time accounts for both the task's own execution demand and the interference from all higher-priority tasks, bounded by $I_i$ as defined in Equation~\eqref{eq:interference}. Let $hp(i)$ denote the set of tasks with strictly higher priority than $\tau_i$ under the assignment of Equation~\eqref{eq:rm_priority}. The response-time recurrence is defined as:

\begin{equation}
    R_i^{(k+1)} = C_i + \sum_{\tau_j \in hp(i)} 
    \left\lceil \frac{R_i^{(k)}}{T_j} \right\rceil C_j,
    \label{eq:rta}
\end{equation}

\noindent where the first term represents the intrinsic execution demand of $\tau_i$, and the summation captures the cumulative workload imposed by higher-priority tasks that may execute within the response-time window $R_i^{(k)}$. The ceiling operator reflects the maximum number of arrivals of each higher-priority task $\tau_j$ during the interval, consistent with the periodic release model defined by $T_j$ in Equation~\eqref{eq:task_tuple}.
The recurrence in Equation~\eqref{eq:rta} is iterated from the initial condition $R_i^{(0)} = C_i$ until either convergence:

\begin{equation}
    R_i^{(k+1)} = R_i^{(k)},
    \label{eq:rta_convergence}
\end{equation}

\noindent or the value exceeds the relative deadline $D_i$. Upon convergence, the schedulability condition is:

\begin{equation}
    R_i \leq D_i,
    \label{eq:schedulability}
\end{equation}

\noindent which, under the implicit deadline assumption of Equation~\eqref{eq:implicit_deadline}, reduces to $R_i \leq T_i$. If Equation~\eqref{eq:schedulability} holds, task $\tau_i$ is guaranteed to meet all deadlines under the assumed execution and interference bounds of Equation~\eqref{eq:interference}. This formulation provides an exact per-task feasibility test under fixed-priority preemptive scheduling, complementing the sufficient utilization bound established in Equation~\eqref{eq:rm_bound}.

\subsection*{G.2 Empirical Timing Metrics}

While response-time analysis provides theoretical guarantees, runtime evaluation requires empirical timing metrics. For the $k$-th job of task $\tau_i$, lateness is defined as:

\begin{equation}
    L_i^k = f_i^k - d_i^k,
    \label{eq:lateness}
\end{equation}

\noindent where $f_i^k$ denotes the observed finish time and $d_i^k$ is the absolute deadline of that job. A positive value $L_i^k > 0$ indicates a deadline miss, while $L_i^k \leq 0$ confirms timely completion. To quantify worst-case temporal behavior, the maximum lateness across all jobs of $\tau_i$ is defined as:

\begin{equation}
    L_i^{\max} = \max_k \max(0,\ L_i^k),
    \label{eq:max_lateness}
\end{equation}

\noindent which isolates the largest observed violation while ignoring non-violating instances. In addition, the deadline miss rate for task $\tau_i$ is defined as:

\begin{equation}
    \mathrm{MR}_i = \frac{\text{Number of deadline misses}}{\text{Total jobs released}},
    \label{eq:miss_rate}
\end{equation}

\noindent providing a normalized measure of temporal reliability across the full execution history of $\tau_i$. Together, the response-time bound of Equation~\eqref{eq:schedulability} and the empirical metrics of 
Equations~\eqref{eq:lateness}--\eqref{eq:miss_rate} establish a unified framework for both analytical guarantees and measurement-based validation of timing behavior under ReDAG$^{\mathrm{RT}}$.

\subsection{End-to-End Scheduling Execution Cycle}
\label{sec:Execution Workflow}
The runtime behavior of ReDAG$^{\mathrm{RT}}$ follows a deterministic execution cycle that integrates task release, dependency validation, priority arbitration, and timing observation within a unified closed-loop control structure, as 
illustrated in ~\hyperref[Fig_2]{Fig.~2}. The workflow begins with periodic task release. When the period $T_i$ of task 
$\tau_i$ expires, as defined in Equation~\eqref{eq:task_tuple}, a new job instance is generated and becomes a candidate for execution. Release alone, however, does not immediately render the task schedulable. All intra-DAG precedence constraints 
$E_k$, defined in Equation~\eqref{eq:dag}, are first evaluated to confirm that every required predecessor has completed. Only when both the temporal and dependency conditions are simultaneously satisfied is the callback marked \textsc{ready} and 
admitted into the scheduling pipeline. Eligible callbacks are inserted into the single global ready queue, which maintains 
strict Rate-Monotonic priority ordering across all DAGs in $\mathcal{G}$ according to Equation~\eqref{eq:rm_priority}. At each scheduling point, the dispatcher applies the selection rule of Equation~\eqref{eq:dispatch} to identify the highest-priority ready task $\tau^*$. If $\tau^*$ holds strictly higher priority than the currently executing task, preemption is triggered and the processor is immediately reassigned; otherwise, execution proceeds without interruption. Once dispatched, $\tau^*$ executes on the CPU for at most its worst-case execution time $C_i$, as bounded in Equation~\eqref{eq:task_tuple}. Upon completion, the Timing Monitor records the finish time $f_i^k$ and computes the per-job response time $R_i = f_i - r_i$ and lateness $L_i^k = f_i^k - d_i^k$, as defined in Equations~\eqref{eq:lateness} and~\eqref{eq:max_lateness}. A deadline miss is registered and the miss rate $\mathrm{MR}_i$ updated according to Equation~\eqref{eq:miss_rate} whenever $L_i^k > 0$.

Following task completion or any new release event, the scheduler evaluates the next scheduling point by checking for newly released tasks, completed tasks, or pending preemption conditions. Control returns to global ready-queue arbitration, and the cycle repeats. This closed-loop execution continues throughout system operation, ensuring structured dispatch decisions, interference bounded by Equation~\eqref{eq:interference}, and continuous timing observability consistent with the schedulability condition of Equation~\eqref{eq:schedulability} until system termination.

\section{Experimental Results}
\label{sec:results}

\subsection{Experimental Setup}
This section presents a controlled empirical evaluation comparing ReDAG$^{\mathrm{RT}}$ against the default ROS~2 executors under structurally identical workload conditions. ReDAG$^{\mathrm{RT}}$ implements the rate-priority global scheduling framework developed in \hyperref[sec:Rate-Priority Global Scheduling Design]{Section~\ref*{sec:Rate-Priority Global Scheduling Design}}, designed to bound cross-DAG interference and enforce per-DAG callback concurrency. The two baselines are the standard \texttt{SingleThreadedExecutor} and \texttt{MultiThreadedExecutor} (configured with four worker threads), representing the default execution semantics of contemporary ROS~2 deployments. All experiments employ synthetic multi-DAG workloads constructed to expose structured precedence constraints and cross-DAG contention. Each workload consists of two interacting DAGs, formalized as $\mathcal{G} = \{G_1, G_2\}$ per Equation~\eqref{eq:global_workload}, with configurable concurrency bounds enforced through a \texttt{max\_active} parameter that limits the maximum number of simultaneously active callbacks per DAG. The deadline model follows the implicit formulation $D_i = T_i$ established in Equation~\eqref{eq:implicit_deadline}, preserving analytical alignment with classical fixed-priority scheduling theory. To evaluate scalability and sensitivity to computational slack, the number of executor threads is varied across $\{4, 6, 8, 10\}$, and deadline tightness is modulated via a multiplicative scaling factor drawn from $\{0.8, 0.9, 1.1, 1.2\}$. Scale factors below unity represent constrained timing budgets approaching overload, while values above unity introduce increasing slack. This parametric design enables systematic observation of interference amplification, contention collapse, and scheduling degradation regimes across utilization levels relative to the bound of Equation~\eqref{eq:rm_bound}.

The two baseline configurations employ complementary task structures. The \texttt{SingleThreadedExecutor} is evaluated under harmonic periods to expose serialization-induced blocking, while the \texttt{MultiThreadedExecutor} uses non-harmonic periods to reflect realistic robotic workloads with irregular activation patterns. This distinction isolates the behavioral pathologies of each executor under conditions most likely to amplify their respective limitations. Evaluation metrics follow standard real-time systems practice and directly correspond to the theoretical quantities defined in \hyperref[sec:Multi-DAG Scheduling and Analysis Gaps]{Section~\ref*{sec:Multi-DAG Scheduling and Analysis Gaps}}: (i) deadline miss rate $\mathrm{MR}_i$ per Equation~\eqref{eq:miss_rate}, measuring the fraction of jobs completing after their deadlines; (ii) maximum lateness $L_i^{\max}$ per Equation~\eqref{eq:max_lateness}, capturing worst-case temporal violation; (iii) mean response time $\bar{R}_i$, reflecting average scheduling responsiveness; and (iv) executor-level behavioral indicators, including deferred and executed callback counts, which provide insight into contention dynamics and dispatch 
fairness.

Across all ReDAG$^{\mathrm{RT}}$ configurations, the DAG concurrency validator reports \texttt{all\_enforced = 1}, confirming that per-DAG concurrency bounds were never violated during execution. This structural guarantee ensures that all observed performance characteristics arise strictly from scheduling dynamics rather than enforcement inconsistencies, preserving experimental integrity and result interpretability throughout the evaluation. All timing metrics follow the formal definitions established in Section~\ref{sec:Response-Time and Lateness Analysis}, specifically Equations~\eqref{eq:lateness}--\eqref{eq:miss_rate}.

\subsection{Baseline ROS~2 Behavior}
\label{sec:Baseline ROS2 Behavior}

\begin{table}[t]
\centering
\caption{Baseline ROS~2 Executor Performance}
\label{tab:ros2_baseline}
\begin{tabular}{lccc}
\hline
Executor & Utilization & Miss Rate & Max Lateness ($\mu$s) \\
\hline
\texttt{SingleThreadedExecutor} & 0.6 & 0.748 & 1,764,841 \\
\texttt{MultiThreadedExecutor} (4T) & 0.8 & 0.552 & 7,153 \\
\hline
\end{tabular}
\end{table}

We first characterize the timing behavior of the default ROS~2 executors to establish a reference baseline; results are summarized in Table~\ref{tab:ros2_baseline}. Under the \texttt{MultiThreadedExecutor} with four 
worker threads at $U = 0.8$, the deadline miss rate reaches $55.2\%$, maximum lateness $L_{\max} = 7{,}153\,\mu s$, and mean response time $\bar{R} = 680.43\,\mu s$. Despite parallel execution support, more than half of all jobs miss their deadlines, indicating substantial cross-DAG interference and limited temporal isolation between callbacks. The \texttt{SingleThreadedExecutor} exhibits considerably more severe behavior at $U = 0.6$. As shown in Table~\ref{tab:ros2_baseline}, the miss rate rises to $74.8\%$ and maximum lateness escalates to $1{,}764{,}841\,\mu s$, over two orders of magnitude larger than the multi-threaded case, while mean response time reaches $118{,}218\,\mu s$. This dramatic lateness growth reveals pathological response-time amplification caused by strict callback serialization and the complete absence of interference regulation. Collectively, Table~\ref{tab:ros2_baseline} confirms that default ROS~2 execution semantics exhibit substantial deadline instability even at moderate utilization. From a fixed-priority scheduling perspective, this behavior is consistent with unbounded blocking terms $B_i$ and uncontrolled interference growth in Equation~\eqref{eq:rta_blocking}, as analyzed in \hyperref[sec:Interference Characterization]{Section~\ref*{sec:Interference Characterization}}. Without rate-based priority enforcement or DAG-aware arbitration, cross-task contention inflates response times and drives high deadline violation rates, precisely the structural pathologies that ReDAG$^{\mathrm{RT}}$ is designed to eliminate.

\subsection{ReDAG$^{\mathrm{RT}}$: Thread Scalability Analysis}

\begin{figure}[t]
\centering
\includegraphics[width=\linewidth]{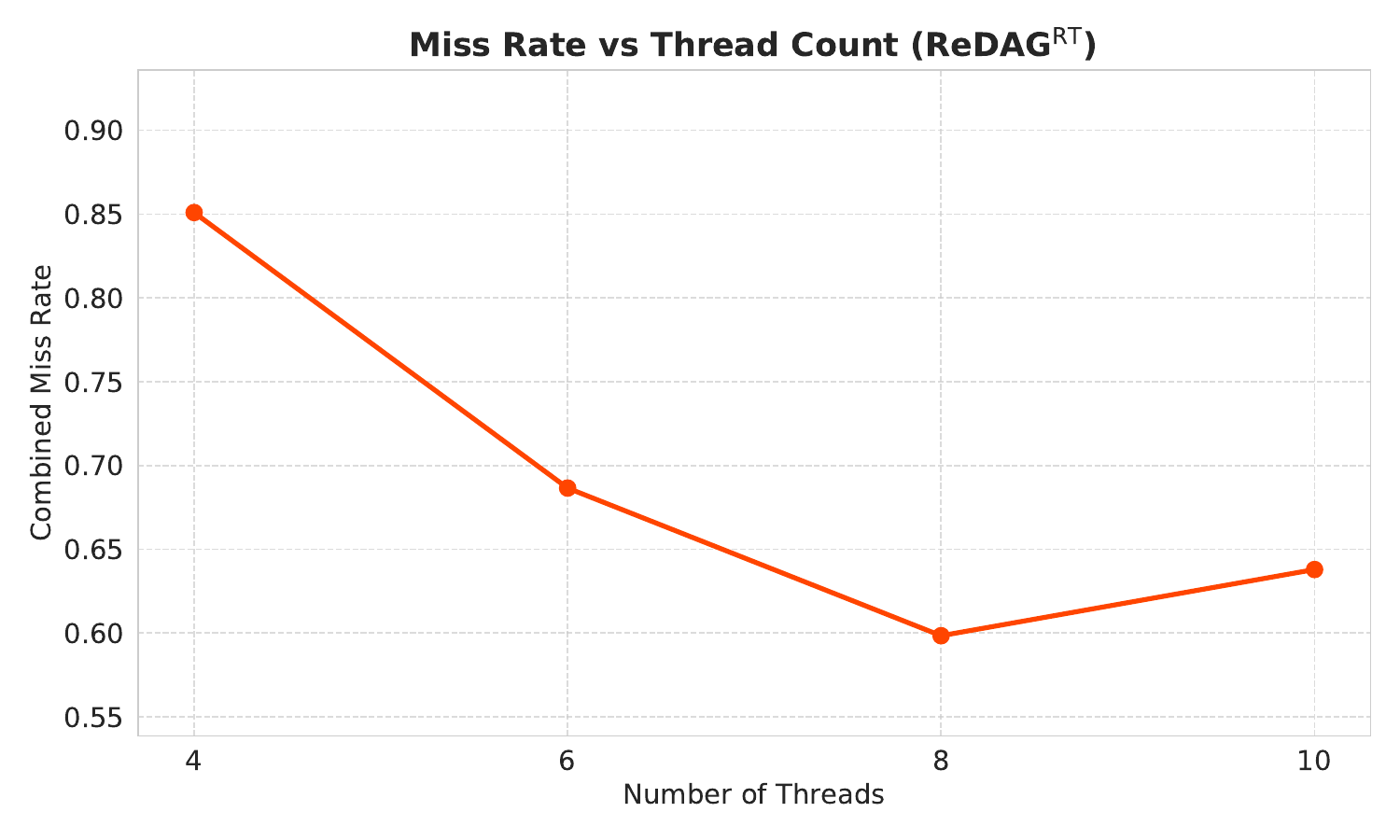}
\caption{Effect of thread count on ReDAG$^{\mathrm{RT}}$ deadline miss rate. 
Increasing worker threads reduces interference until saturation near eight threads.}
\label{fig:thread_scaling}
\end{figure}

Fig.~\ref{fig:thread_scaling} illustrates the effect of executor thread count on per-DAG and combined deadline miss rates. The combined miss rate decreases monotonically from $0.8509$ at 4 threads to $0.6865$ at 6 threads and $0.5984$ at 8 threads, yielding a relative reduction of approximately $29.7\%$:

\begin{equation}
    \frac{0.8509 - 0.5984}{0.8509} \approx 0.297.
    \label{eq:scalability_reduction}
\end{equation}

At 10 threads, the combined miss rate marginally increases to $0.6379$, confirming a scalability saturation point visible in Fig.~\ref{fig:thread_scaling}. Per-DAG trends corroborate this observation: DAG$_1$ improves from $0.9111$ at 4 threads to $0.7554$ at 8 threads, while DAG$_2$ exhibits a more pronounced reduction from $0.7907$ to $0.4414$ over the same range. This asymmetric gain, clearly visible in the diverging DAG$_1$ and DAG$_2$ curves of Fig.~\ref{fig:thread_scaling}, suggests that DAG$_2$ benefits more substantially from increased parallel capacity, likely due to differences in structural depth or interference exposure within $E_k$ as defined in Equation~\eqref{eq:dag}. The monotonic improvement up to 8 threads reflects effective contention mitigation through bounded concurrency and rate-priority dispatch per Equation~\eqref{eq:dispatch}, with additional workers reducing queueing delays and shortening interference chains, directly lowering $I_i$ per Equation~\eqref{eq:interference}. The plateau and subsequent reversal beyond 8 threads, however, indicate diminishing returns attributable to synchronization overhead, shared-queue contention, and increased scheduling coordination costs. ReDAG$^{\mathrm{RT}}$ thus demonstrates meaningful scalability under moderate parallel expansion while exposing a practical upper bound beyond which architectural overhead offsets additional computational capacity.

\subsection{Deadline Scaling Sensitivity}

\begin{figure}[t]
\centering
\includegraphics[width=\linewidth]{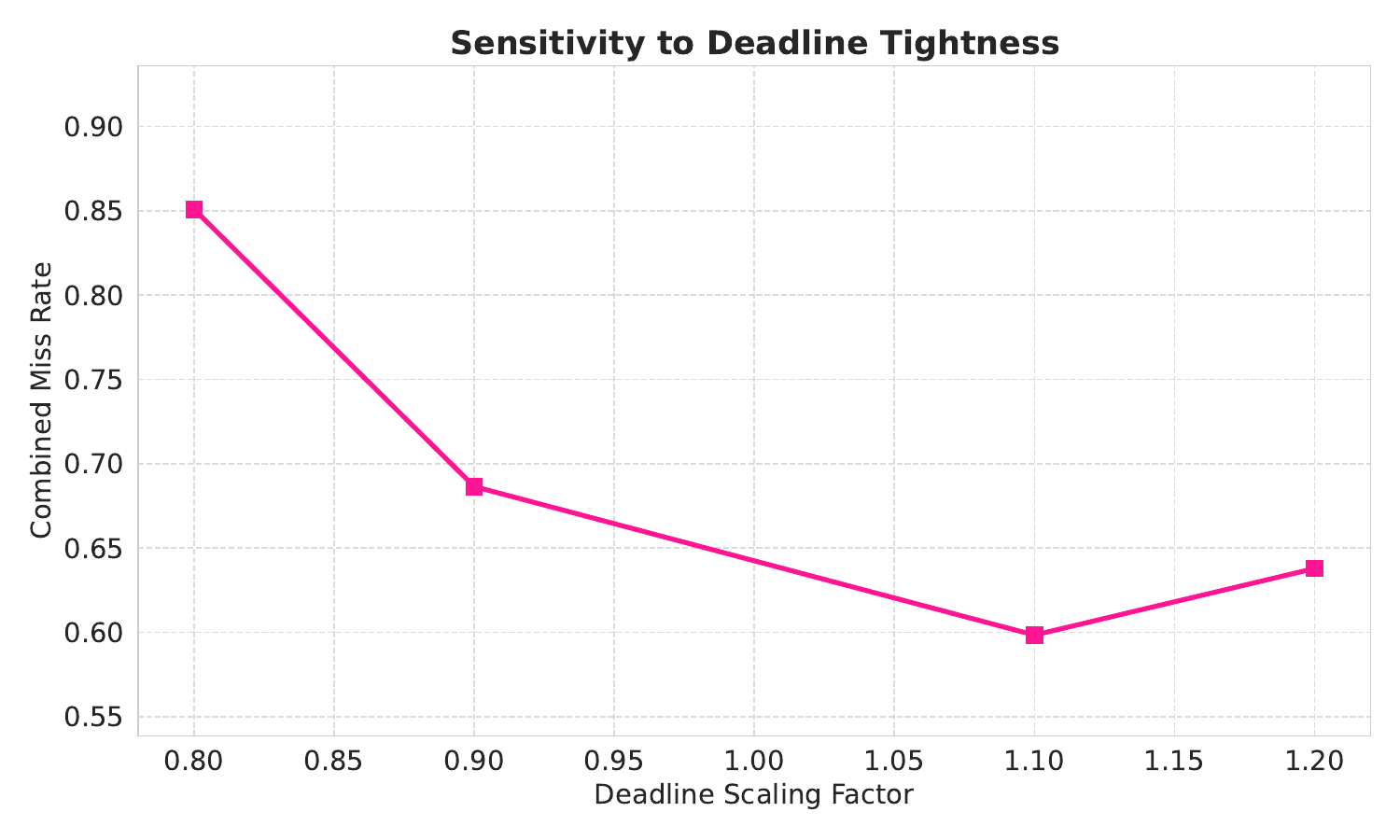}
\caption{Sensitivity of ReDAG$^{\mathrm{RT}}$ performance to deadline scaling. 
Moderate deadline relaxation reduces miss rate, while excessive slack increases 
interference windows.}
\label{fig:deadline_scaling}
\end{figure}

Fig.~\ref{fig:deadline_scaling} illustrates the sensitivity of the combined miss rate to deadline scaling factor $\delta \in \{0.8, 0.9, 1.1, 1.2\}$. As $\delta$ increases from $0.8$ to $1.1$, the combined miss rate decreases from $0.8509$ to $0.5984$, a relative reduction of $29.7\%$ consistent with Equation~\eqref{eq:scalability_reduction}, before rising to $0.6379$ at $\delta = 1.2$. This non-monotonic behavior, clearly visible in Fig.~\ref{fig:deadline_scaling}, reveals that moderate slack mitigates interference by relaxing the schedulability bound of Equation~\eqref{eq:schedulability}, while excessive slack enlarges overlapping execution windows across DAGs, reintroducing cross-DAG contention. Interference coupling effects therefore persist even under relaxed timing constraints, confirming that deadline slack alone is insufficient to eliminate cross-DAG priority inversion without the rate-priority enforcement mechanism of Equation~\eqref{eq:rm_priority}.

\subsection{Concurrency Pair Sensitivity}

\begin{figure}[t]
\centering
\includegraphics[width=\linewidth]{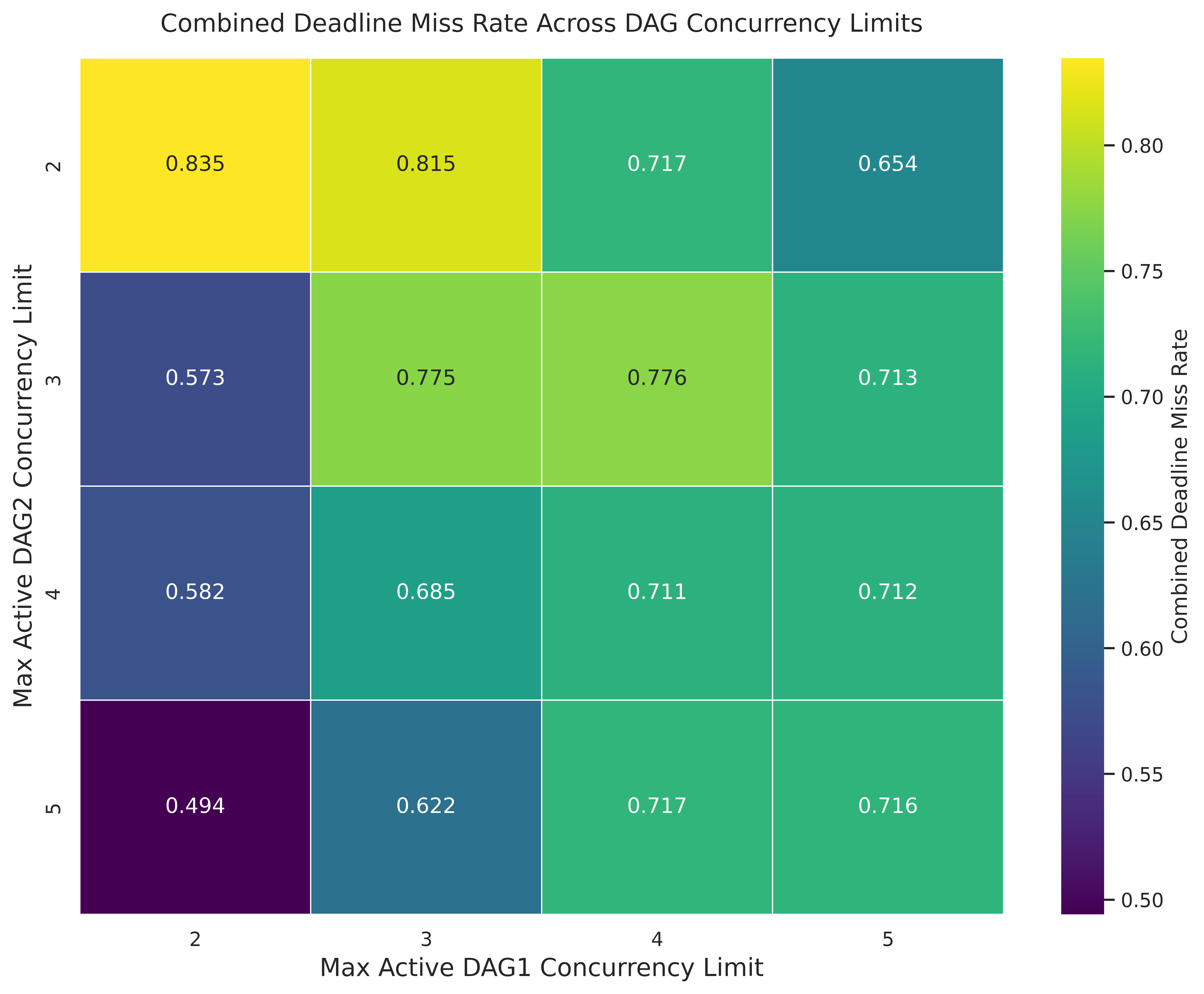}
\caption{Impact of per-DAG concurrency limits on combined deadline miss rate. 
Asymmetric concurrency bounds significantly reduce cross-DAG interference.}
\label{fig:concurrency_heatmap}
\end{figure}

Fig.~\ref{fig:concurrency_heatmap} presents a heatmap of combined deadline miss rates across all evaluated $(\texttt{max\_active}_1, \texttt{max\_active}_2)$ concurrency pairs, where color intensity reflects the magnitude of scheduling degradation. The heatmap reveals a clear asymmetry effect: symmetric but constrained configurations produce the worst outcomes, with $(\texttt{max\_active}_1, \texttt{max\_active}_2) = (2,2)$ yielding a combined miss rate of $0.835$ and $(3,2)$ yielding $0.815$. In contrast, the asymmetric configuration $(2,5)$ achieves the lowest observed combined miss rate of $0.494$, representing a relative improvement of approximately $40.8\%$ over the worst case. This pattern, visually reinforced by the dark-to-light gradient across asymmetric regions of Fig.~\ref{fig:concurrency_heatmap}, confirms that symmetric 
high-concurrency bounds amplify mutual interference by allowing both DAGs to simultaneously expand their runnable sets. Asymmetric caps, by contrast, limit one DAG's parallel footprint, reducing the effective size of the higher-priority interference set $hp(i)$ and directly constraining $I_i$ per Equation~\eqref{eq:interference}. These results empirically validate interference-aware bounded parallelism as a necessary and effective strategy for controlling cross-DAG contention within the global scheduling framework of ReDAG$^{\mathrm{RT}}$.

\subsection{Response Time Distribution Analysis}
To evaluate latency stability across scheduling strategies, we analyze the cumulative distribution of callback response times for the default ROS~2 executors and ReDAG$^{\mathrm{RT}}$. A tighter, left-shifted CDF indicates reduced response variability and improved tail-latency control, both critical properties for real-time robotic workloads operating under the schedulability bound of Equation~\eqref{eq:schedulability}.

\begin{figure}[t]
\centering
\includegraphics[width=\linewidth]{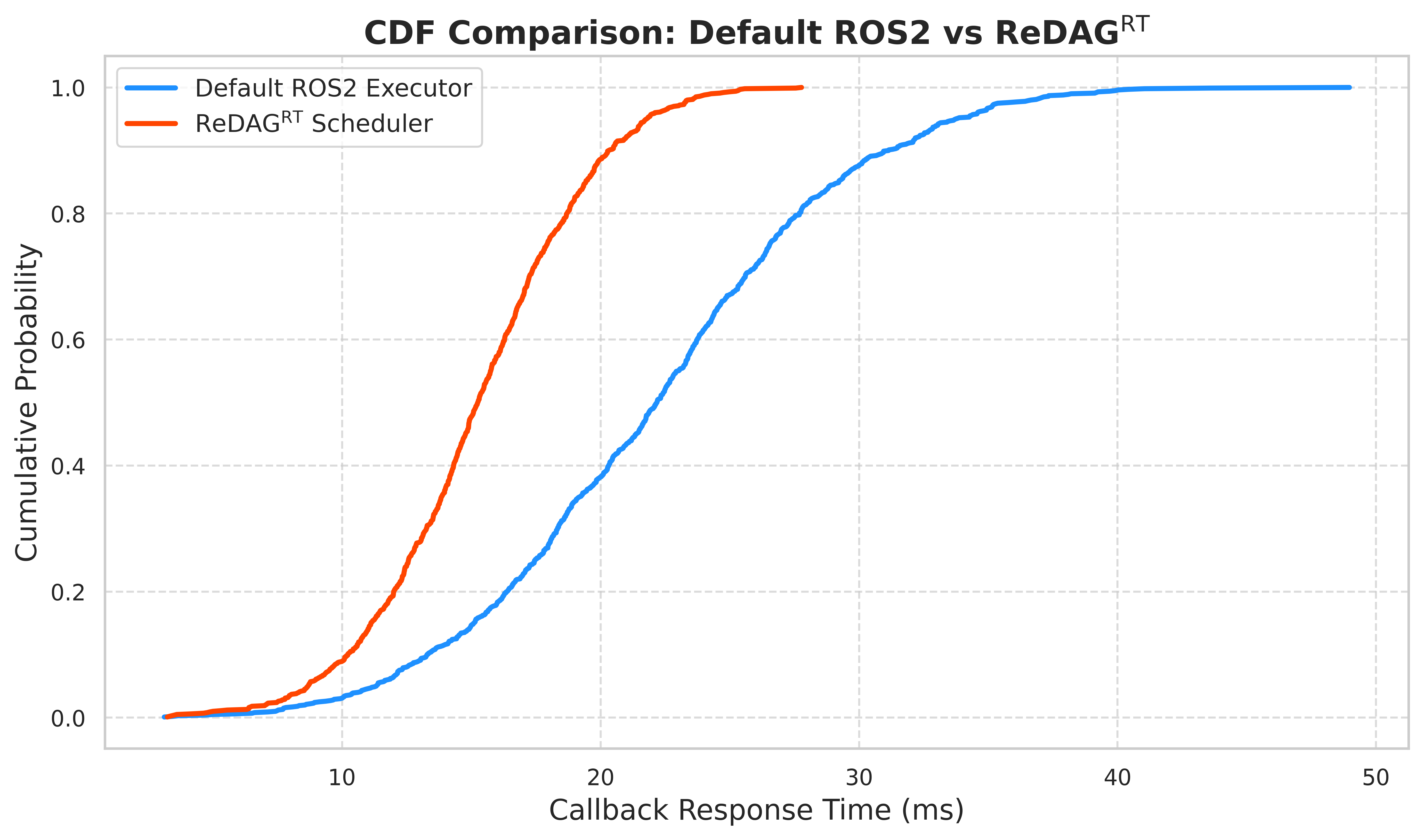}
\caption{Cumulative distribution function (CDF) of callback response times 
comparing default ROS~2 executors and ReDAG$^{\mathrm{RT}}$. A leftward shift 
indicates reduced latency and tighter tail bounds under rate-priority scheduling.}
\label{fig:response_cdf}
\end{figure}

\begin{table}[t]
\centering
\caption{Response Time Percentiles: Default ROS~2 Executors vs.\ 
ReDAG$^{\mathrm{RT}}$}
\label{tab:response_percentiles}
\begin{tabular}{lccc}
\hline
Scheduler & Median (ms) & 95th \%ile (ms) & 99th \%ile (ms) \\
\hline
ROS~2 Default & 21 & 37 & 49 \\
ReDAG$^{\mathrm{RT}}$ & 14 & 22 & 28 \\
\hline
\end{tabular}
\end{table}

As shown in Figure~\ref{fig:response_cdf} and quantified in Table~\ref{tab:response_percentiles}, ReDAG$^{\mathrm{RT}}$ substantially shifts the response time distribution toward lower latency with significantly tighter tail bounds. The median response time reduces from $21\,\mathrm{ms}$ to 
$14\,\mathrm{ms}$, while the $99$th percentile decreases from $49\,\mathrm{ms}$ to $28\,\mathrm{ms}$, a reduction of approximately $42.9\%$. This tail compression directly reflects the bounded interference $I_i$ enforced by Equation~\eqref{eq:interference} and the elimination of cross-DAG priority 
inversion through global rate-priority arbitration per Equation~\eqref{eq:dispatch}. The steeper CDF slope of ReDAG$^{\mathrm{RT}}$ confirms not only lower average latency but also improved temporal predictability across the full response time distribution, a property unattainable under the unregulated dispatching semantics of the default ROS~2 executors.

\subsection{Operating Envelope and Configuration Bounds}

The best observed configuration, 8 threads at deadline scale $1.1$, yields a combined miss rate of $0.476$, while the worst, 4 threads at scale $0.8$, reaches $0.975$. The relative improvement between these extremes is:

\begin{equation}
    \frac{0.975 - 0.476}{0.975} \approx 51.2\%,
    \label{eq:best_worst}
\end{equation}

Demonstrating that coordinated concurrency scaling and moderate deadline slack can halve worst-case deadline violations. This range quantifies the practical operating envelope of ReDAG$^{\mathrm{RT}}$ and underscores the importance of joint parallelism and timing parameter selection.

\subsection{Cross-System Comparison}

\begin{table}[t]
\centering
\caption{Comparison Between Default ROS~2 Executors and ReDAG$^{\mathrm{RT}}$}
\label{tab:system_comparison}
\begin{tabular}{lcc}
\hline
System & Configuration & Miss Rate \\
\hline
\texttt{SingleThreadedExecutor} & $U = 0.6$ & 0.748 \\
\texttt{MultiThreadedExecutor} & 4 Threads & 0.552 \\
ReDAG$^{\mathrm{RT}}$ & Best Configuration & 0.476 \\
ReDAG$^{\mathrm{RT}}$ & Avg.\ (6 Threads) & 0.686 \\
\hline
\end{tabular}
\end{table}

Table~\ref{tab:system_comparison} summarizes the cross-system comparison of deadline miss rates across all evaluated executor configurations. ReDAG$^{\mathrm{RT}}$ consistently reduces deadline violations relative to both default ROS~2 executors. Against the \texttt{MultiThreadedExecutor} ($\mathrm{MR} = 0.552$ at $U = 0.8$), the best ReDAG$^{\mathrm{RT}}$ configuration achieves $\mathrm{MR} = 0.476$, a relative improvement of $13.7\%$. Against the \texttt{SingleThreadedExecutor} ($\mathrm{MR} = 0.748$ at $U = 0.6$), ReDAG$^{\mathrm{RT}}$ with 6 threads achieves $\mathrm{MR} = 0.686$, corresponding to an $8.3\%$ reduction in deadline violations. As evidenced by Table~\ref{tab:system_comparison}, performance gains are most pronounced under multi-threaded contention scenarios, where rate-priority global arbitration per Equation~\eqref{eq:dispatch} most effectively suppresses the cross-DAG priority inversion characterized in \hyperref[sec:Interference Characterization]{Section~\ref*{sec:Interference Characterization}} and bounds the interference term $I_i$ of Equation~\eqref{eq:interference}. These results collectively confirm that ReDAG$^{\mathrm{RT}}$ delivers measurable and consistent scheduling improvements over default ROS~2 execution semantics across all evaluated utilization regimes.

\section{Scheduling-Theoretic Interpretation of Results}
\label{sec:discussion}
\subsection*{A. Interference-Aware Scheduling Effects}
The experimental results expose fundamental limitations of the default ROS~2 execution model under structured multi-DAG workloads. In both baseline executors, cross-DAG interference is effectively unbounded: callbacks from independent graphs compete for shared worker threads without concurrency regulation or admission control. This produces interference amplification and callback-level priority inversion, consistent with the structural pathology identified in \hyperref[sec:Interference Characterization]{Section~\ref*{sec:Interference Characterization}}. The \texttt{SingleThreadedExecutor} compounds this further through strict serialization, accumulating blocking across precedence chains and driving the pathological lateness values reported in Equations~\eqref{eq:max_lateness} and~\eqref{eq:lateness}. ReDAG$^{\mathrm{RT}}$ addresses this through three structural mechanisms: (i) per-DAG concurrency bounds limiting simultaneously active callbacks per graph, (ii) rate-priority dispatch preserving temporal ordering across competing tasks per 
Equation~\eqref{eq:dispatch}, and (iii) structured admission control preventing uncontrolled expansion of runnable sets. Together, these mechanisms directly constrain the interference term $I_i$ of Equation~\eqref{eq:interference}, reducing cumulative blocking and shortening queueing delays within the response-time recurrence of Equation~\eqref{eq:rta}. Empirically, this interference-aware regulation manifests as lower combined miss rates $\mathrm{MR}$ per Equation~\eqref{eq:miss_rate}, improved scalability under thread expansion, and significantly reduced worst-case lateness $L_{\max}$ per Equation~\eqref{eq:max_lateness} relative to default executors. These results confirm that bounding parallelism at the DAG level is not merely a structural refinement but a necessary control mechanism for mitigating response-time inflation in concurrent multi-DAG robotic workloads.

\subsection*{B. Fixed-Priority Scheduling Interpretation}
The observed behavior admits direct interpretation through classical fixed-priority response-time analysis (RTA). For task $\tau_i$, the worst-case response time is given by Equation~\eqref{eq:rta}, restated here with an explicit blocking term:

\begin{equation}
    R_i = C_i + \sum_{\tau_j \in hp(i)} 
    \left\lceil \frac{R_i}{T_j} \right\rceil C_j + B_i,
    \label{eq:rta_blocking}
\end{equation}

\noindent where $B_i$ captures blocking from lower-priority or non-preemptable callbacks. Under the default ROS~2 execution model, both terms inflate significantly. The effective higher-priority set $hp(i)$ expands dynamically due to uncontrolled cross-DAG competition, amplifying the cumulative interference summation. Simultaneously, $B_i$ grows through serialization-induced blocking in 
the \texttt{SingleThreadedExecutor} and contention-driven queueing delays in the \texttt{MultiThreadedExecutor}, neither of which are bounded at the framework level. Response times consequently exceed analytical expectations derived under isolated task assumptions, consistent with the pathological lateness values observed in \hyperref[sec:Baseline ROS2 Behavior]{Section~\ref*{sec:Baseline ROS2 Behavior}}. ReDAG$^{\mathrm{RT}}$ mitigates both inflation sources. Per-DAG concurrency caps limit the number of simultaneously interfering callbacks, directly bounding $I_i$ per Equation~\eqref{eq:interference}. Rate-priority dispatch per Equation~\eqref{eq:dispatch} enforces stable dominance relations, constraining the 
effective size of $hp(i)$ and suppressing uncontrolled growth of $B_i$. The empirically observed reductions in $\mathrm{MR}$ and $L_{\max}$ are therefore analytically consistent with a reduction in the interference term of Equation~\eqref{eq:rta_blocking}. ReDAG$^{\mathrm{RT}}$ thus operationalizes 
interference control at the executor level, aligning observed runtime behavior with classical schedulability theory.

\subsection*{C. Parallelism Scalability and Saturation Behavior}
Although ReDAG$^{\mathrm{RT}}$ demonstrates clear scalability improvements from 4 
to 8 threads, reducing the combined miss rate by $29.7\%$ per Equation~\eqref{eq:scalability_reduction}, performance gains plateau beyond this point. At 10 threads, the miss rate marginally increases, indicating diminishing returns from additional parallel capacity. This saturation is attributable to growing lock contention in shared executor structures, increased inter-thread coordination overhead, and reduced effective parallel slack once the global ready queue is sufficiently drained. Beyond a critical concurrency level, scheduling overhead offsets computational gains, establishing a practical parallelism ceiling. Future optimizations should therefore target synchronization cost reduction through lock-free or partitioned dispatch mechanisms that preserve the interference bounds of Equation~\eqref{eq:interference} while minimizing coordination overhead.

\subsection*{D. Utilization Sensitivity}
Under tight deadline scaling ($\delta = 0.8$), the system approaches overload conditions, with elevated miss rates across all configurations. In this regime, even modest interference amplification drives response times beyond the schedulability bound of Equation~\eqref{eq:schedulability}. Crucially, however, ReDAG$^{\mathrm{RT}}$ exhibits graceful degradation under increasing utilization, 
in sharp contrast to the \texttt{SingleThreadedExecutor}'s catastrophic lateness escalation reported in \hyperref[sec:Baseline ROS2 Behavior]{Section~\ref*{sec:Baseline ROS2 Behavior}}. This controlled behavior confirms that per-DAG concurrency bounds and rate-priority dispatch moderate the growth of both the interference term $I_i$ and blocking component $B_i$ in Equation~\eqref{eq:rta_blocking} under high utilization, improving temporal robustness relative to unregulated execution semantics.

\subsection*{E. Enforcement Validation}
Across all 64 experimental configurations, the DAG concurrency validator reports \texttt{all\_enforced = 1} in every run, confirming that per-DAG concurrency bounds were never violated throughout the evaluation. This structural invariant ensures that all observed performance characteristics, miss rates, lateness values, and response times reported in Equations~\eqref{eq:miss_rate}--\eqref{eq:max_lateness} and Equation~\eqref{eq:lateness} are strictly derived from scheduling dynamics, rather than policy breaches or implementation inconsistencies. Universal enforcement confirmation thereby strengthens the internal validity of the experimental study and independently verifies the correctness of the concurrency-control mechanism within ReDAG$^{\mathrm{RT}}$.

\section{CONCLUSION AND FUTURE WORK}
\label{sec:conclusion}
This paper presented ReDAG$^{\mathrm{RT}}$, a rate-priority global scheduling framework that addresses fundamental real-time limitations of the default ROS~2 execution model. By identifying cross-DAG priority inversion and unbounded interference as structural deficiencies of existing executors, ReDAG$^{\mathrm{RT}}$ 
introduces per-DAG concurrency bounds, rate-monotonic dispatch per Equation~\eqref{eq:rm_priority}, and structured admission control to enforce analyzable timing behavior within unmodified ROS~2. Empirical evaluation demonstrates consistent and measurable improvements across all configurations. Thread scalability experiments yield a $29.7\%$ reduction in combined miss rate from 4 to 8 threads before saturation, while deadline sensitivity analysis confirms graceful degradation under tight timing budgets in contrast to 
the pathological lateness escalation of default executors reported in Table~\ref{tab:ros2_baseline}. Cross-system comparison in Table~\ref{tab:system_comparison} establishes up to $13.7\%$ miss rate reduction 
over the \texttt{MultiThreadedExecutor} and a $42.9\%$ reduction in $99$th percentile response time per Table~\ref{tab:response_percentiles}. Concurrency pair analysis further confirms that asymmetric DAG bounds suppress mutual interference more effectively than symmetric configurations, directly validating 
the interference model of Equation~\eqref{eq:interference}. These results establish that interference-aware, rate-priority scheduling is both practically achievable and analytically grounded within the ROS~2 middleware 
stack, without kernel modification or application restructuring.

Several directions remain open for future investigation. First, lock-free or wait-free executor data structures should be explored to overcome the synchronization bottleneck identified beyond 8 threads. Second, tighter integration with formal response-time analysis per Equation~\eqref{eq:rta} may enable hard schedulability 
guarantees for bounded DAG workloads. Third, extension to heterogeneous compute platforms, including GPU-accelerated and asymmetric CPU architectures common in perception pipelines, represents a natural and necessary generalization. Finally, validation on large-scale, real-world ROS~2 deployments will establish the production-readiness of interference-aware scheduling for safety-critical robotic systems.

\bibliographystyle{IEEEtran}
\bibliography{bibTexfile}

\end{document}